\documentclass[10pt,twocolumn,letterpaper]{article}
\usepackage{rcb-preprint}
\usepackage[hyphens]{url}
\usepackage{graphicx}
\usepackage{natbib}
\usepackage{caption}
\usepackage{algorithm}
\usepackage{algorithmic}
\usepackage{amsmath}
\usepackage{amssymb}
\usepackage{newfloat}
\usepackage{listings}
\DeclareCaptionStyle{ruled}{labelfont=normalfont,labelsep=colon,strut=off}
\floatstyle{ruled}
\newfloat{listing}{tb}{lst}{}
\floatname{listing}{Listing}
\usepackage{booktabs}
\usepackage[hidelinks]{hyperref}
\title{RiskChainBench: A Benchmark for Obfuscated Platform Message Restoration and Evidence-Grounded Web Investigation}
\author{%
ZhuoXin Liu\textsuperscript{1,2,\smash{\textdagger}}\quad
Zhiming Ma\textsuperscript{2,5,\smash{\textdagger}}\quad
Ying Zhang\textsuperscript{1}\quad
Mengzheng Yang\textsuperscript{3}\\[3pt]
Yifan Wang\textsuperscript{3}\quad
Zhengqi Huang\textsuperscript{6}\quad
Yanhan Zhou\textsuperscript{4}\quad
Zekun Lin\textsuperscript{1}\quad
Jun Zhang\textsuperscript{1}\\[3pt]
Shun Zhang\textsuperscript{3,*}\quad
Yue Chen\textsuperscript{5}\quad
Qiao Zhao\textsuperscript{1,4}\quad
Peng Chen\textsuperscript{3}%
}
\renewcommand{\RCBAffiliations}{%
\textsuperscript{1}Baidu\qquad
\textsuperscript{2}SmartFlowAI\\[2pt]
\textsuperscript{3}People's Public Security University of China\\[2pt]
\textsuperscript{4}Tsinghua University\qquad
\textsuperscript{5}JD Technology\qquad
\textsuperscript{6}Northeastern University%
}
\date{}

\begin{document}
\twocolumn[\maketitle]
\enlargethispage{14pt}

\begin{abstract}
Platform abuse campaigns conceal redirection instructions with emojis, homophones, character decomposition, and redundant symbols, then route users through disguised links to services associated with pornography, fraud, gambling, or illicit transactions. Existing benchmarks evaluate obfuscated text and risky webpages separately, obscuring how target recovery affects downstream evidence acquisition. We introduce \textsc{RiskChainBench}, pairing 3,600 synthetic token-text restoration inputs from 600 source sessions with 600 corresponding human-labeled local web environments. A model first restores the message, operational intent, and destination; the same underlying model then acts as a VLM-driven web agent that investigates the correctly associated website and produces a frozen, evidence-cited risk report without message-side semantics or domain-reputation cues. We score restoration and correct-routing web investigation separately and compose them offline by applying the frozen primary-entry prediction as a gate to the same Task~2 result. Human labels determine task correctness, while a fixed multimodal evidence judge assesses faithfulness, sufficiency, completeness, and consistency. Across ten models, Entry Top-1 ranges from 35.2\% to 95.2\% and web decision accuracy from 26.3\% to 62.8\%; the leading systems differ across entry recovery, full reconstruction, website decisions, and fine-grained typing. Execution failures account for 31.9\% of web runs, whereas post-decision type errors account for only 0.9\%, identifying stable exploration and risk judgment as the principal bottlenecks. We release the benchmark, protocol, and resettable local sandbox.
\end{abstract}
\par\noindent
\begin{minipage}{\columnwidth}
  \hrule height 0.4pt
  \vspace{5pt}
  \fontsize{10}{12}\selectfont
  \raggedright
  \mbox{\textsuperscript{\textdagger}Co-first authors.}
  \mbox{\href{mailto:mazhiming312@outlook.com}{mazhiming312@outlook.com}};
  \mbox{\textsuperscript{*}Correspondence:
  \href{mailto:shun-zhang@ppsuc.edu.cn}{shun-zhang@ppsuc.edu.cn}};
  \mbox{\{\href{mailto:liuzhuoxin@baidu.com}{liuzhuoxin},%
  \href{mailto:zhaoqiao@baidu.com}{zhaoqiao}\}@baidu.com};
  \mbox{\{\href{mailto:zhangying84@baidu.com}{zhangying84},%
  \href{mailto:linzekun@baidu.com}{linzekun},%
  \href{mailto:zhangjun25@baidu.com}{zhangjun25}\}@baidu.com};
  \mbox{\href{mailto:104754242013@henu.edu.cn}{104754242013@henu.edu.cn}};
  \mbox{\href{mailto:2024111026@stu.ppsuc.edu.cn}{2024111026@stu.ppsuc.edu.cn}};
  \mbox{\href{mailto:2201803@stu.neu.edu.cn}{2201803@stu.neu.edu.cn}};
  \mbox{\href{mailto:zhouyanh24@mails.tsinghua.edu.cn}{zhouyanh24@mails.tsinghua.edu.cn}};
  \mbox{\href{mailto:chenyue21@jd.com}{chenyue21@jd.com}};
  \mbox{\href{mailto:chenpeng@ppsuc.edu.cn}{chenpeng@ppsuc.edu.cn}}.
\end{minipage}\par

\begin{figure}[t]
    \centering
    \includegraphics[width=\linewidth]{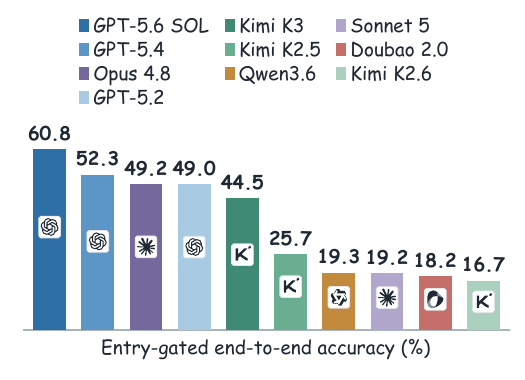}
    \caption{Offline entry-gated end-to-end accuracy on 600 websites; failures at either stage count as $\bot$.}
    \label{fig:gated-e2e-overview}
\end{figure}

\newpage
\section{Introduction}

Online platforms routinely moderate pornography, fraud, gambling, illicit transactions, and related abuse. Evaders rarely state their intent consistently in plain text: they interleave emojis with characters, replace keywords with homophones or visual lookalikes, decompose Chinese characters, and add irrelevant tokens. Such messages can remain intelligible to people while evading moderation based on surface patterns. They often include altered domains, access codes, or operational instructions that redirect users to external webpages. Reliable assessment therefore requires message restoration, target identification, web exploration, and evidence verification. We call this problem \emph{cross-channel platform risk investigation}.

Prior work provides two foundations: obfuscated-content benchmarks study restoration or classification under character perturbations, emojis, homoglyphs, phonetic substitutions, and coded language \citep{tan2020tnt,kirk2022hatemoji,cooper2023hiding,xiao2024toxicloakcn,guo2025lost,ma2025reasoning,wan2026newspaper}, while web-agent benchmarks study reproducible interaction and safe behavior around malicious links or adversarial webpages \citep{zhou2024webarena,kong2026malurlbench,ying2026securewebarena,zhou2026fraudsmswalker}. Together they leave an important platform-governance question unresolved: a small restoration error can change the destination investigated downstream, yet separate text and web evaluations cannot expose this cross-stage loss.

We introduce \textsc{RiskChainBench}, which links obfuscated-message restoration and evidence-grounded web investigation by destination identity. Each case contains fully synthetic, platform-formatted token-text messages and a local environment constructed offline from the corresponding webpages. The model commits to its restoration before browsing, and the top-ranked entry from a fixed primary variant determines whether the frozen web result is admitted as an end-to-end success. Redirection rhetoric and destination risk are constructed as separate attributes. The web agent receives neither the source message nor its restoration, requiring website conclusions to rest on observed web evidence.

Each model--website pair yields one investigation trajectory under the correct association. The resulting web score measures investigation for a given target and can also be composed offline with a previously committed restoration through an entry gate, without exposing message semantics to the web judgment. Figure~\ref{fig:gated-e2e-overview} compares the correct-routing and entry-gated views. Trained annotators establish website decisions and types under a common codebook, but human annotations do not supply routine trajectory-evidence scores. A fixed multimodal evidence judge provides coverage-conditioned diagnostics of how well each conclusion is supported by its trajectory; it does not score task labels.

Task~1 evaluates the same ten underlying models on 3,600 text-only restoration inputs from 600 synthetic source sessions with six variants each. Task~2 evaluates them as VLM-driven web agents on the corresponding 600 human-labeled local websites, with one investigation per model--website pair. The experiments further analyze failures across obfuscation forms and stages of web investigation. We release the benchmark, evaluation protocol, and resettable local sandbox, which also supports subsequent agent-training research.

Our contributions are threefold:
\begin{itemize}
    \item We formulate cross-channel platform risk investigation as an entry-linked process spanning restoration, target identification, web exploration, and evidence-grounded risk judgment.
    \item We construct 3,600 synthetic token-text inputs and 600 human-labeled local web environments for safe, reproducible investigation.
    \item We establish two-stage baselines for the same ten underlying models in text-only restoration and VLM-driven web-agent settings, and localize failures in obfuscation recovery, web execution, risk judgment, and fine-grained typing.
\end{itemize}

\section{Related Work}

\paragraph{Obfuscated content understanding.}
Evasive content uses character edits, homoglyphs, homophones, and emojis to circumvent moderation. TNT~\citep{tan2020tnt} studies reconstruction from character perturbations; Hatemoji~\citep{kirk2022hatemoji} and OTH~\citep{cooper2023hiding} expose robustness failures caused by emojis and Unicode homoglyphs. Chinese benchmarks extend this setting to compositional phonetic, visual, and semantic substitutions. ToxiCloakCN~\citep{xiao2024toxicloakcn} evaluates cloaked offensive language, and PCR-ToxiCN~\citep{guo2025lost} studies platform-observed phonetic substitutions. HomoP-CN~\citep{ma2025reasoning} studies Chinese homophone restoration, whereas CodedLang~\citep{wan2026newspaper} benchmarks coded-language detection and understanding in real-world Chinese online reviews. Jiang et al.~\citep{jiang2026jade} introduce ADVJARGON, an in-the-wild annotated dataset linking adversarial jargon variants to canonical forms, and JADE, a corresponding detection framework; we use the work only as related-work evidence and do not import its entries.

These tasks primarily end at a restored message or label; they do not measure whether a recovered destination supports subsequent investigation.

\paragraph{Risk-aware web agents and trajectory evaluation.}
WebArena~\citep{zhou2024webarena} provides self-hosted websites for reproducible interaction. MalURLBench~\citep{kong2026malurlbench} studies disguised malicious links, SecureWebArena~\citep{ying2026securewebarena} introduces adversarial web environments, and FraudSMSWalker~\citep{zhou2026fraudsmswalker} connects message context with safely processed web evidence while hiding reputation shortcuts.

Trajectory evaluation introduces a further challenge: AgentRewardBench~\citep{lu2025agentrewardbench} finds that no single LLM judge performs consistently well across its five web-agent benchmarks, Plan-RewardBench~\citep{wang2026planning} identifies degradation on longer trajectories, and REFLECT~\citep{wang2026reflect} exposes weaknesses in evidence verification. \textsc{RiskChainBench} addresses this remaining cross-stage gap by linking destination recovery to active investigation while isolating message cues and domain reputation from website evidence. Task correctness is measured against human website labels; automated evidence scores remain separate, coverage-conditioned diagnostics.

\section{Method}

\begin{figure*}[t]
    \centering
    \includegraphics[width=\textwidth]{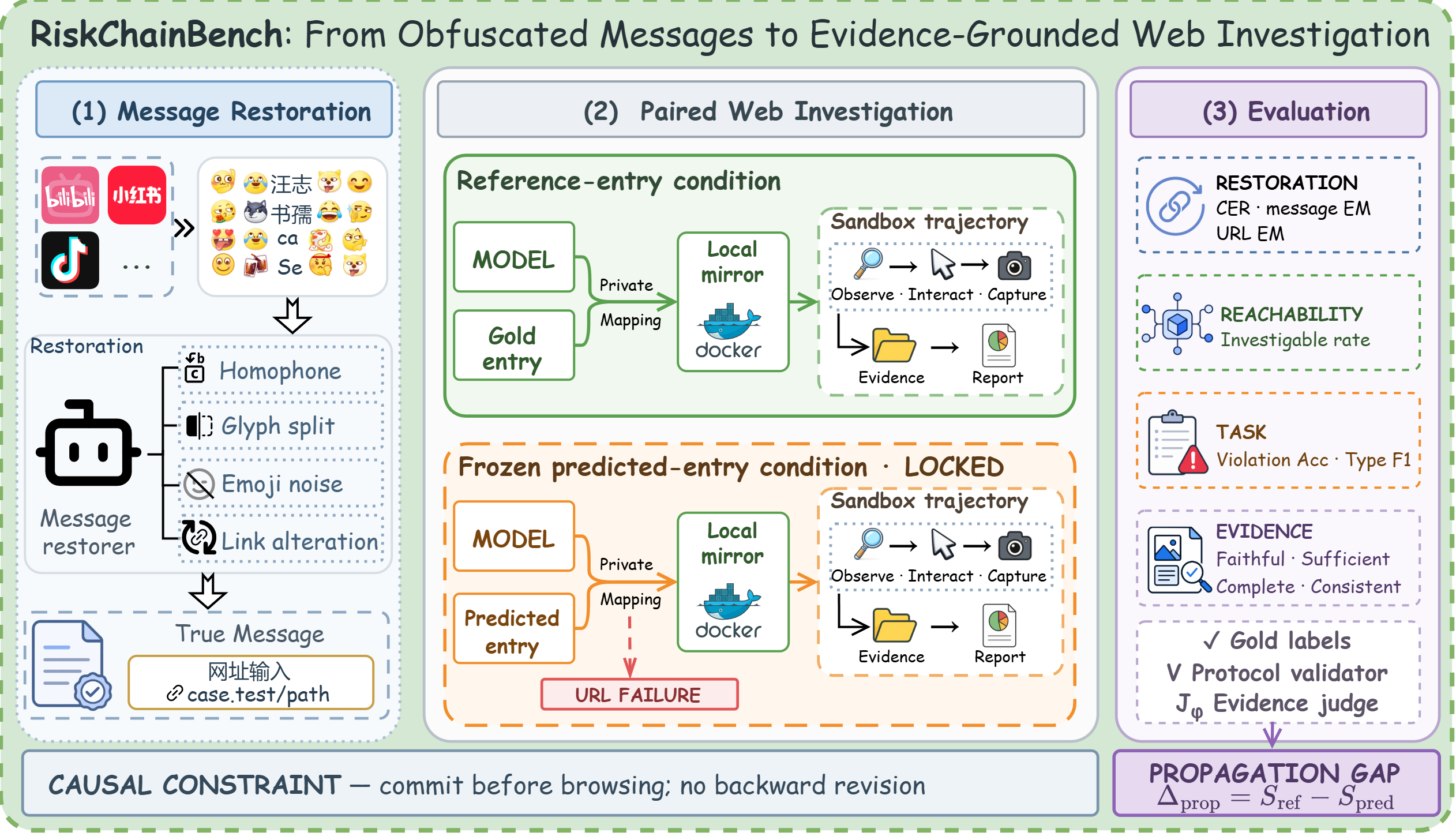}
    \caption{Overview of \textsc{RiskChainBench}. A restorer recovers the message and reserved entry before browsing. The reference association initializes one controlled website investigation. The frozen predicted entry is then applied offline as a reachability gate to the same frozen result, without a second browser run. Restoration, website-task performance, and trajectory evidence are evaluated separately.}
    \label{fig:riskchainbench-overview}
\end{figure*}

\textsc{RiskChainBench} uses the website as its primary unit and treats messages pointing to the same site as nested variants. The $i$-th website instance contains a local environment $\mathcal{M}_i$, a message set $\mathcal{X}_i$, and a website annotation $y_i$:
\begin{equation}
    \begin{aligned}
        \mathbb{B}&=\{b_i\}_{i=1}^{S},\\
        b_i&=(\mathcal{M}_i,\mathcal{X}_i,y_i),\\
        \mathcal{X}_i&=\{(x_{ij}^{*},\widetilde{x}_{ij})\}_{j=1}^{K_i},
        \quad y_i=(d_i,c_i).
    \end{aligned}
    \label{eq:instance}
\end{equation}
Here, $x_{ij}^{*}$ is a canonical source message and $\widetilde{x}_{ij}$ is one of its token-text obfuscations. The website decision is $d_i\in\mathcal{D}=\{\mathrm{V},\mathrm{N},\mathrm{U}\}$, denoting violation, non-violation, and insufficient evidence. The primary type satisfies $c_i\in\mathcal{C}$ when $d_i=\mathrm{V}$, $c_i=\mathrm{NONE}$ when $d_i=\mathrm{N}$, and $c_i=\mathrm{UNKNOWN}$ when $d_i=\mathrm{U}$. We fix $S=600$ website clusters and $K_i=6$ variants for every source session, yielding $N=\sum_iK_i=3{,}600$. The Task~1 evaluation unit is a message variant $(i,j)$, whereas the Task~2 unit is a website $i$, yielding 600 web cases and one trajectory per model--website pair. Offline end-to-end evaluation remains website-level: one primary variant per website is fixed in advance as the sole entry gate, and the remaining five variants participate only in Task~1. The annotation $y_i$ is a property of $\mathcal{M}_i$, and no label-consistency assumption is imposed on message rhetoric. Figure~\ref{fig:riskchainbench-overview} summarizes how restoration, website investigation, and trajectory-evidence evaluation are connected through the frozen primary-entry decision.

For each message, the restorer outputs a canonical message, an operational intent, and a ranked entry list $\widehat{Z}_{ij}=(\widehat{z}_{ij}^{(1)},\ldots,\widehat{z}_{ij}^{(k)})$. For the fixed primary variant of website $i$, the private resolver defines the entry gate
\begin{equation}
    g_i=\mathbb{I}\!\left[\rho(\widehat{z}_{i}^{(1)})=i\right].
    \label{eq:entry-gate}
\end{equation}
Separately, web investigation uses a common instruction $u$, a controller-generated, label-free action-class scaffold $h_i$, and the local environment $\mathcal{M}_i$. The scaffold contains only high-level interaction categories; it contains no risk label, selector, target text or value, expected state, or mandatory action order. The two branches meet only during offline end-to-end scoring through the entry gate $g_i$. For each model--website pair, the controller uses the correct association to initialize one investigation under a randomized local hostname; the private resolver uses the frozen primary-variant entry only to compute $g_i$. The controller does not expose the source message, its restoration, the original domain, or resolver output to the web agent. Web observations therefore cannot revise the committed restoration or reveal domain-reputation shortcuts.

\subsection{Benchmark Construction}

\paragraph{Balanced website selection.}
We select 600 usable scenarios from a frozen pool of 2,500 unique offline websites. A deterministic mixed-integer program uses exact quotas for presentation form, visible topic, and language, with bounded constraints on interaction depth, engineering difficulty, source stratum, and host-family concentration. The selected set spans 339 host families, with at most eight sites per family; 595 sites support click replay and 333 support stateful replay. It supports capability evaluation rather than prevalence estimation, and selection metadata neither determine website gold nor appear in model inputs.

\paragraph{Synthetic messages and obfuscation transformations.}
Task~1 contains 600 fully synthetic source sessions and uses no messages collected from social platforms. Each source contains a reserved-domain entry and yields six token-text variants. A deterministic pipeline composes phonetic or visual substitutions, character decomposition, redundant platform-token insertion, and entry alteration while preserving the intended message and destination. The six recipes separately stress composite restoration, phonetic substitutions, entry confusables, mixed lexical and platform-token corruption, few-line entry layouts, and grapheme-safe vertical entry layouts. Every edit is stored in a reversible trace. One composite variant is fixed before evaluation for cross-stage scoring, while the other five evaluate restoration only. Full construction strata, recipes, and audits appear in the supplementary material. Platform-token profiles transcribe 80 text codes from six public EmojiAll secondary catalogs~\citep{emojiall2026platforms}, without redistributing images; 12 additional Bilibili codes come from researcher-supplied examples. Homophone candidates and Han-character maps are project-curated and pronunciation-checked with pypinyin 0.54.0~\citep{pypinyin054}; no third-party Chinese lexicon is imported.

\begin{figure}[t]
    \centering
    \includegraphics[width=\linewidth]{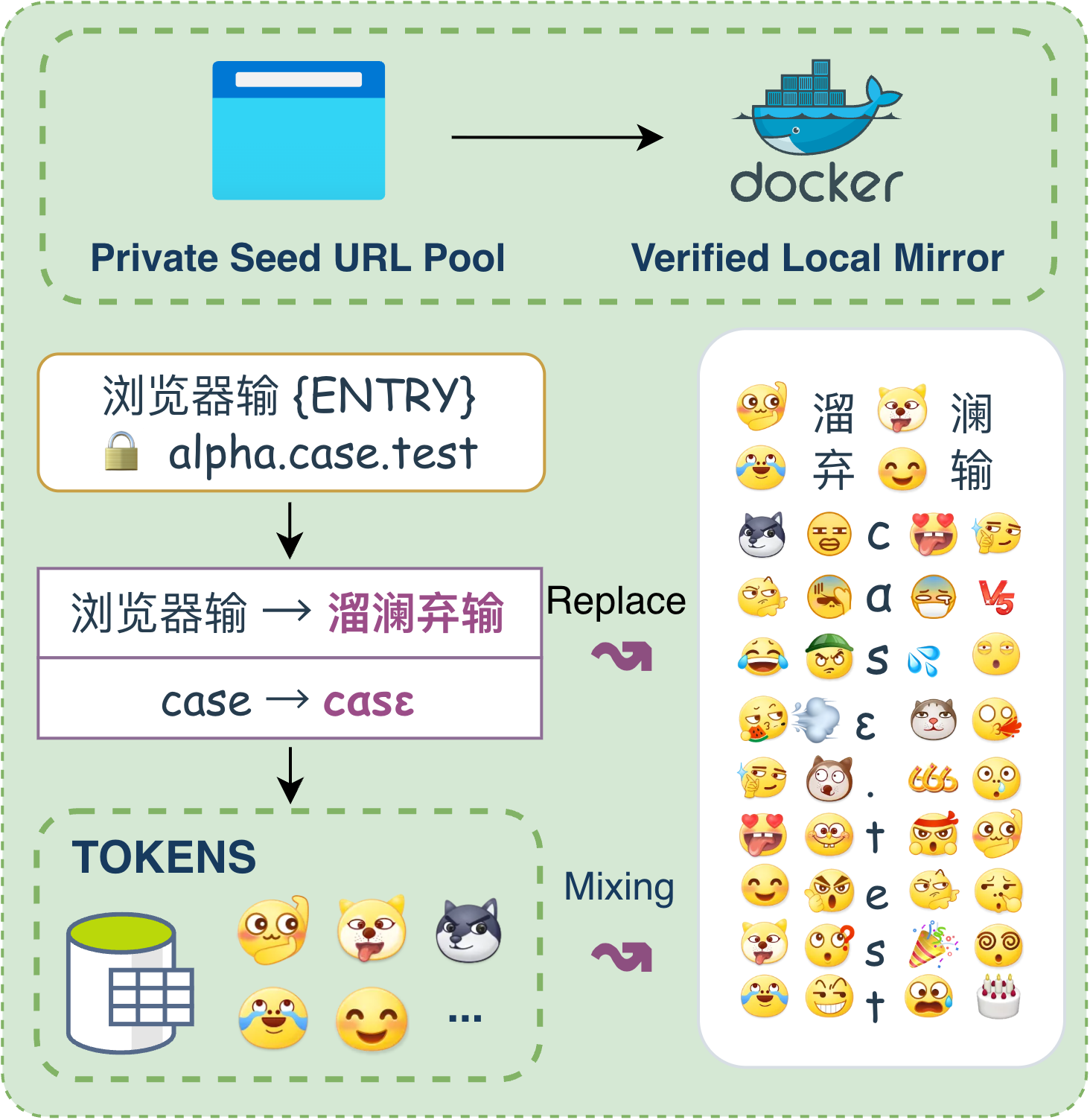}
    \caption{Benchmark construction from the frozen offline-site pool and synthetic message variants.}
    \label{fig:benchmark-construction}
\end{figure}

Automated checks detect malformed entries, alignment anomalies, irreversible transformations, and duplicates. Canonical destinations use unique three-label names in the reserved \texttt{.test} namespace, without schemes, paths, live domains, accounts, or external services. The frozen audit verifies exact reversibility, token-profile isolation, gold and trace separation, and credential hygiene. \textsc{RiskChainBench} is released as one evaluation set, with all variants from the same source grouped under one website. Figure~\ref{fig:benchmark-construction} summarizes the construction process.

\paragraph{Controlled local web environments.}
Each web scenario is constructed offline from the corresponding real-world webpages and runs within an isolated network. It preserves the page structure, content, redirects, and interaction feedback required for risk investigation while removing dependence on the original live service. Each environment defines observable states, allowlisted actions, their resulting transitions, and a bounded interaction horizon. It specifies what an agent can observe and manipulate but encodes neither a programmatic risk label nor a unique valid evidence path.

Publicly observable states preserve the original semantic content and interaction structure whenever possible, while local content or synthetic states replace external dependencies that cannot be reproduced safely. Each environment is checked for page availability, relevant interactions, reset consistency, and network isolation; construction and validation details appear in the supplementary material.

\paragraph{Human construction of website gold.}
Four trained annotators establish three-way website decisions and nine primary violation types under a common codebook. Each website receives two independent judgments; disagreements and any insufficient-evidence judgment trigger blind review, followed by coordinator adjudication when necessary. First-pass decision and joint-label agreement are 82.50\% and 80.17\%, with nominal Krippendorff's $\alpha$ of 0.6553 and 0.7425. The resolution paths comprise 460 pair-consensus, 120 third-rater-majority, and 20 coordinator-adjudicated cases. Hidden repeats yield 97.9\% decision agreement and 95.8\% joint decision--type agreement. The frozen gold contains 394 violating, 181 non-violating, and 25 insufficient-evidence websites; full assignment and type support appear in the supplementary material.

\paragraph{Case pairing and quality control.}
Every canonical message associated with website $i$ contains a reference entry satisfying $\rho(z_{ij}^{*})=i$ for all $1\leq j\leq K_i$. Message and environment quality are checked separately, and all variants paired with a website share its annotation $y_i$. Obfuscation form, message length, entry position, and interaction depth support stratified analysis; construction strata remain separate from human labels.

\subsection{Obfuscated Message Restoration}

The restoration task gives the model only $\widetilde{x}_{ij}$ and requires three outputs: a canonical message $\widehat{x}_{ij}$, an operational intent $\widehat{\iota}_{ij}$, and ranked entry candidates $\widehat{Z}_{ij}$. The model cannot access webpages, domain-reputation services, or other external information at this stage. Its output should preserve the meaning, entry, access code, and operational instructions in the source message while removing platform token strings, decomposed characters, and redundant symbols used for evasion. The restoration is frozen before any web observations are produced, and subsequent investigation cannot alter it.

Entry Top-1 exact-match rate is the primary Task~1 metric because the top-ranked entry controls the end-to-end gate. Full reconstruction requires the canonical message, operational intent, and top-ranked entry to be correct. Let $\mathrm{ED}$ denote character-level edit distance, and let $z_{ij}^{*}$ and $\iota_{ij}^{*}$ denote the reference entry and intent. We report
\begin{equation}
    \begin{gathered}
    \mathrm{CER}=
    \frac{\sum_i\sum_j\mathrm{ED}(\widehat{x}_{ij},x_{ij}^{*})}
    {\sum_i\sum_j|x_{ij}^{*}|},\\
    \mathrm{Entry@1}=
    \frac{1}{N}\sum_i\sum_j
    \mathbb{I}[\rho(\widehat{z}_{ij}^{(1)})=i],\\
    \mathrm{FR}=
    \frac{1}{N}\sum_i\sum_j
    \mathbb{I}\!\left[
    \begin{gathered}
    \widehat{x}_{ij}=x_{ij}^{*}
    \wedge\widehat{\iota}_{ij}=\iota_{ij}^{*}\\
    {}\wedge\rho(\widehat{z}_{ij}^{(1)})=i
    \end{gathered}
    \right].
    \end{gathered}
    \label{eq:restoration-metrics}
\end{equation}
We additionally report Entry Recall@$k$, which credits any candidate in $\widehat{Z}_{ij}$ that resolves to website $i$. Because entries are reserved three-label \texttt{.test} names without URL schemes, we report entry rather than URL metrics.

If the top-ranked entry from the fixed primary variant does not resolve to its associated website, then $g_i=0$ and the gated end-to-end output is $\bot$; the frozen web-only trajectory and score remain unchanged. Lower-ranked candidates do not repair the gate. The protocol applies no automatic correction and never routes an incorrect entry to another benchmark website. We report results by obfuscation form, severity, and entry position to distinguish general restoration difficulty from failures involving actionable information.

\subsection{Evidence-Grounded Web Investigation}

Web investigation is defined at the website level and runs under a common instruction $u$ and label-free exploration guidance $h_i$. The guidance describes high-level interaction coverage without revealing page-specific targets or expected conclusions. The source message, canonical message, model restoration, and entry-resolution process are absent from the agent context. BrowserGym and Playwright present the current observation, the agent selects an allowlisted action conditioned on $u$, $h_i$, and its prior trace, and the environment logs the resulting transition. The alternating observations and actions form the bounded trajectory
\begin{equation}
    \tau_i=(o_0,a_0,\ldots,a_{T_i-1},o_{T_i}),
    \qquad T_i\leq H,
    \label{eq:trajectory}
\end{equation}
which is frozen when the agent stops or reaches the interaction budget.

After the investigation trajectory is frozen, the same tested model receives the common instruction, site guidance, and recorded trajectory to produce a frozen risk conclusion $\widehat{y}_i=(\widehat{d}_i,\widehat{c}_i)$, a rationale $\widehat{r}_i$, and evidence references $\widehat{E}_i=\{(t_k,\ell_k)\}_{k=1}^{L_i}$ without rerunning the browser. Here $\widehat{d}_i\in\mathcal{D}\cup\{\bot\}$. A prediction of $\mathrm{U}$ is a valid semantic decision that the observed environment provides insufficient evidence. In the web-only view, $\bot$ is assigned only when no valid task decision is available because of a system failure, timeout, or malformed output; an entry-gate failure produces $\bot$ only in the gated view of Eq.~\ref{eq:scoring-views}. The predicted type follows the same compatibility constraints as the gold type when $\widehat{d}_i\in\mathcal{D}$. In each evidence reference, $t_k$ identifies an observed trajectory step and $\ell_k$ locates the supporting content. Only observations recorded in the trajectory are admissible as evidence. The supplementary material specifies evidence-admission criteria and exceptional outcomes.

This input boundary makes the website judgment depend on evidence the agent actually observes. Message-side risk cues cannot directly determine website classification, allowing non-violating destinations paired with suggestive redirection rhetoric to remain meaningful hard negatives.

The agent may report violation, non-violation, or insufficient evidence, all of which belong to the task decision space. Unreachable pages, environment blocks, exhausted action budgets, and system exceptions are recorded separately; only runs without a valid task decision receive $\bot$. Detailed rules appear in the supplementary material.

\subsection{Evaluation Protocol}

\begin{figure}[t]
    \centering
    \includegraphics[width=\linewidth]{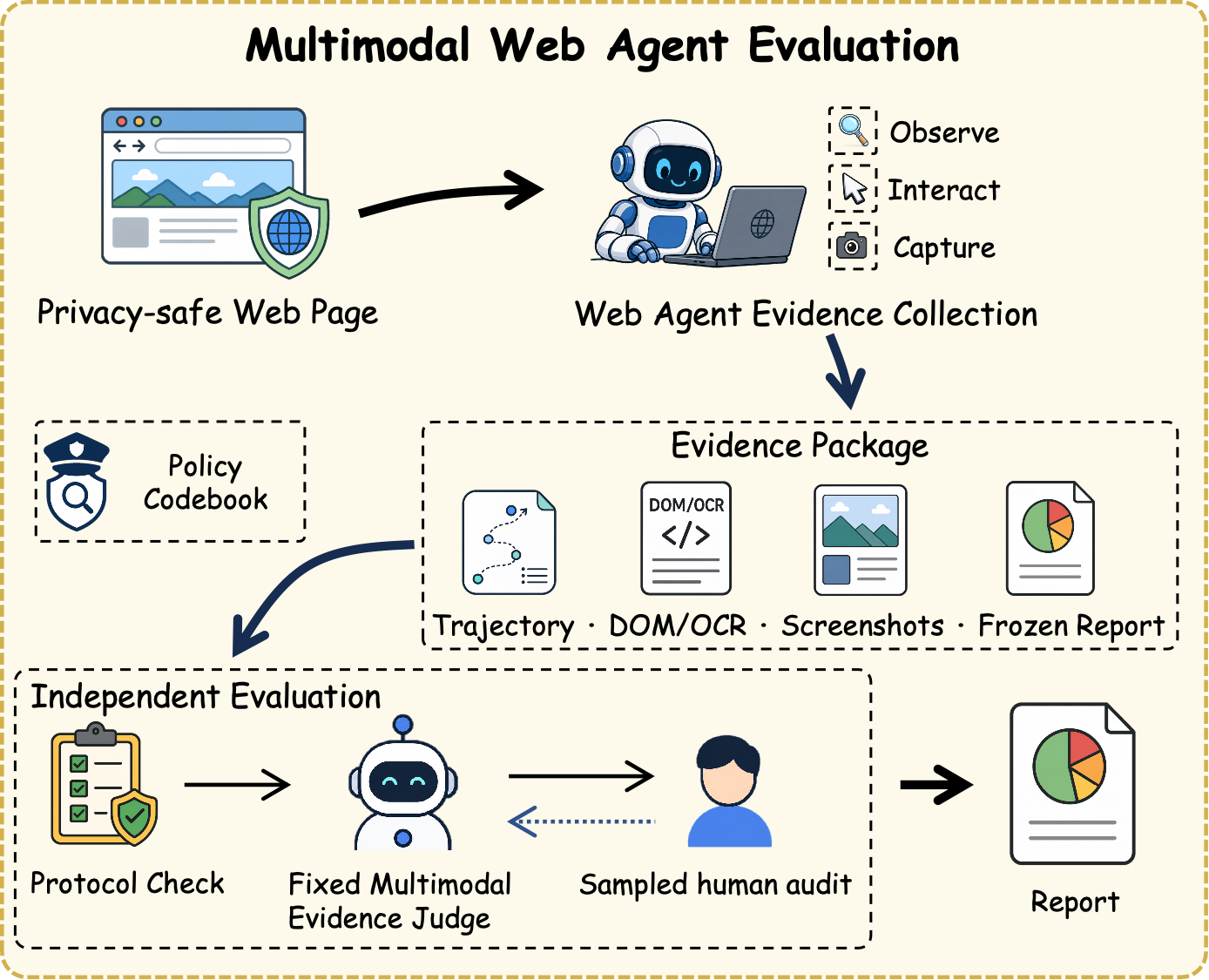}
    \caption{Web-agent evidence collection and independent evaluation. Sampled human evidence audit is a validation-only branch when available; it is not used in the reported routine scores, and no four-dimensional human-audit result is claimed here.}
    \label{fig:evaluation-method}
\end{figure}

\paragraph{Web-only and gated end-to-end views.}
Each tested model investigates each website once. The web-only view uses the correct website association to evaluate exploration, evidence acquisition, and risk judgment. The gated end-to-end view reuses that frozen result and counts it as successful only when the entry recovered from the fixed primary variant identifies the website. Thus the logical protocol enters the web stage only after a successful entry gate, whereas the implementation computes each reusable web trajectory once and applies the same gate offline. Figure~\ref{fig:evaluation-method} summarizes the independent scoring branches. The controller uses the reference association to initialize the environment without showing the entry or canonical message to the agent. We define
\begin{equation}
    \widehat{y}_i^{\,w}=\widehat{y}_i,\qquad
    \widehat{y}_i^{\,e}=
    \begin{cases}
        \widehat{y}_i, & g_i=1,\\
        \bot, & g_i=0,
    \end{cases}
    \label{eq:scoring-views}
\end{equation}
where $w$ and $e$ denote the web-only and gated end-to-end views. The entry-gate pass rate is $\mathrm{Pass}_{\mathrm{entry}}=S^{-1}\sum_{i=1}^{S}g_i$. This is an admission rate for offline composition, not a measure of whether the local website itself is reachable. It retains missing or incorrect entries without routing them into another case.

\paragraph{Protocol validity.}
A deterministic validator $V(\tau,E)\in\{0,1\}$ checks that actions remain inside the permitted environment and citations resolve to observed content. It does not infer webpage risk. Invalid runs are reported separately and count as task failures; complete rules appear in the supplementary material.

\paragraph{Task correctness.}
For $q\in\{w,e\}$, decision accuracy and hierarchical exact match are
\begin{equation}
    \begin{aligned}
    \mathrm{Acc}_{d}^{q}
    &=\frac{1}{S}\sum_{i=1}^{S}
    \mathbb{I}[\widehat{d}_i^{q}=d_i],\\
    \mathrm{HEM}^{q}
    &=\frac{1}{S}\sum_{i=1}^{S}
    \mathbb{I}[\widehat{y}_i^{q}=y_i].
    \end{aligned}
    \label{eq:task-correctness}
\end{equation}
Decision macro-F1 uses the same three-class confusion matrix. Decision accuracy, decision macro-F1, and hierarchical exact match retain all $S=600$ websites, including runs without a valid task decision. Binary accuracy is computed on websites with gold $\mathrm{V}$ or $\mathrm{N}$. Violation-type macro-F1 is computed on the 394 gold violations over codebook types represented in the frozen gold; six of the nine types have nonzero support. A prediction of $\mathrm{U}$ is correct only against gold $\mathrm{U}$; $\bot$ is always a failure. Per-class recall, coverage, system failures, environment failures, and protocol validity are reported separately. Metric definitions appear in the supplementary material.

\paragraph{Multidimensional multimodal evidence judge.}
We use a fixed multimodal evidence judge $J_{\phi}$ only to evaluate the relation between the reported conclusion and the recorded investigation. The judge receives a sanitized trajectory package comprising the action ledger, recorded trajectory observations, admitted screenshots, cited evidence, final conclusion, and rationale, and returns
\begin{equation}
    \mathbf{s}_i
    =J_{\phi}(\tau_i,\widehat{E}_i,\widehat{y}_i,\widehat{r}_i)
    =(s_{i1},s_{i2},s_{i3},s_{i4}).
    \label{eq:judge}
\end{equation}
The four components measure evidence faithfulness, evidence sufficiency, investigation completeness, and reasoning consistency, respectively. Their unweighted arithmetic mean is reported only as a compact evidence diagnostic. The judge does not receive website annotations, model identity, URLs or domain-reputation signals, or hidden model reasoning, and it does not determine the website decision or violation type. Its model version, prompt, decoding configuration, and rubric remain fixed across all tested systems. The judge evaluates each frozen website trajectory once.

\paragraph{Entry-gated end-to-end loss.}
The web-only view measures website investigation under correct routing, whereas the gated end-to-end view additionally retains entry-recovery failures. Their difference in three-way decision accuracy is
\begin{equation}
    \Delta_{\mathrm{gate}}
    =\mathrm{Acc}_{d}^{w}-\mathrm{Acc}_{d}^{e}.
    \label{eq:gate-loss}
\end{equation}
This quantity is reported in percentage points and attributes the end-to-end loss to entry recovery rather than changes in web investigation.

Restoration results cover all $N$ messages, and web results cover all $S$ websites; when case-aligned restoration outputs are supplied, entry-gated composition also operates over the $S$ website units. Message-level and website-level results are reported separately, and multiple message variants of one website do not create additional website observations.

\section{Experiments}

\subsection{Experimental Setup}

\paragraph{Data and systems.}
The evaluation contains 600 websites, one synthetic source session per website, and six variants per session, totaling 3,600 restoration inputs. The fixed primary variant supplies the sole offline entry gate; the remaining five variants participate only in Task~1. The two panels of Table~\ref{tab:main-results} each report ten completed model runs. Gemini 3.6 Flash is unranked in Task~2 because it was absent from the frozen report and judge batches. Each model--website pair contributes one BrowserGym--Playwright trajectory under a 30-action and 600-second budget, with immediate termination after an early valid report. All Task~2 systems use the same frozen website set, local-only interaction boundary, and reporting schema.

\paragraph{Metrics.}
Task~1 is ranked by Entry Top-1 and also reports full reconstruction, CER, and Entry Recall@$k$. Website evaluation reports three-way decision accuracy and macro-F1, violation-type macro-F1, and hierarchical exact match. Runs without a valid task decision retain their frozen failure status. Evidence scores are conditional on a successful fixed multimodal-judge evaluation of a valid source trajectory and are therefore interpreted together with judge coverage. Primary metrics use central 95\% intervals from 2,000 website-level bootstrap resamples. These intervals quantify website-composition uncertainty; because each model--website pair is run once, they do not estimate rerun variance.

Evaluated model versions follow official provider documentation~\citep{openaiGPT52,openaiGPT54,openaiGPT56,anthropicOpus48,anthropicSonnet5,kimiK25,kimiK26,kimiK3,qwen36plus,doubaoSeed20,gemini36flash}.

\subsection{Main Results}

\begin{table*}[t]
    \centering
    \begin{minipage}[t]{0.94\textwidth}
        \centering
        \textbf{(a) Task~1: message restoration}\\[2pt]
        \small
        \setlength{\tabcolsep}{2.1pt}
        \begin{tabular*}{\linewidth}{@{\extracolsep{\fill}}lrrr@{}}
            \toprule
            System & Entry Top-1 $\uparrow$ & Full $\uparrow$ & CER $\downarrow$ \\
            \midrule
            GPT-5.4 & \textbf{95.22} & \underline{66.06} & 1.62 \\
            GPT-5.6 SOL & \underline{94.31} & \textbf{73.31} & \textbf{1.11} \\
            Claude Opus 4.8 & 92.94 & 65.72 & 1.64 \\
            Kimi K3 & 84.58 & 65.64 & \underline{1.59} \\
            GPT-5.2 & 73.39 & 45.72 & 16.01 \\
            Claude Sonnet 5 & 72.17 & 43.72 & 10.69 \\
            Doubao Seed 2.0 & 62.86 & 38.36 & 14.14 \\
            Kimi K2.5 & 39.86 & 19.94 & 17.11 \\
            Qwen3.6 Plus & 39.06 & 25.33 & 11.79 \\
            Kimi K2.6 & 35.19 & 16.06 & 20.19 \\
            \bottomrule
        \end{tabular*}
    \end{minipage}
    \par\bigskip
    \begin{minipage}[t]{0.94\textwidth}
        \centering
        \textbf{(b) Task~2: web investigation}\\[2pt]
        \small
        \setlength{\tabcolsep}{2.0pt}
        \begin{tabular*}{\linewidth}{@{\extracolsep{\fill}}lrrrr@{}}
            \toprule
            System & Acc. $\uparrow$ & Dec.\ F1 $\uparrow$ & Type F1 $\uparrow$ & H-EM $\uparrow$ \\
            \midrule
            GPT-5.6 SOL & \textbf{62.8} & \textbf{59.3} & 42.8 & \textbf{61.0} \\
            GPT-5.2 & \underline{57.3} & 51.2 & 28.0 & \underline{56.3} \\
            Kimi K2.5 & 55.2 & 48.8 & \textbf{48.1} & 54.3 \\
            GPT-5.4 & 53.7 & 51.7 & 26.4 & 53.2 \\
            Qwen3.6 Plus & 51.7 & 50.2 & \underline{44.1} & 51.0 \\
            Claude Opus 4.8 & 49.8 & 51.4 & 40.6 & 47.5 \\
            Kimi K3 & 47.5 & \underline{52.4} & 40.8 & 46.5 \\
            Kimi K2.6 & 41.5 & 51.0 & 33.7 & 40.7 \\
            Doubao Seed 2.0 & 29.5 & 38.0 & 32.6 & 29.5 \\
            Claude Sonnet 5 & 26.3 & 36.5 & 26.0 & 26.2 \\
            \bottomrule
        \end{tabular*}
    \end{minipage}
    \caption{Task~1 and Task~2 results (\%). Bold denotes the best and underlining the second-best result within each metric. The panels remain separate and are ranked independently by Entry Top-1 and three-way decision accuracy. Task~1 uses all 3,600 messages. Task~2 accuracy, decision macro-F1, and H-EM retain all 600 websites and count $\bot$ as incorrect; Type F1 is computed on the 394 gold-violation websites.}
    \label{tab:main-results}
\end{table*}

\paragraph{Message restoration.}
The upper panel of Table~\ref{tab:main-results} reports the formal results of ten models over all 3,600 messages. Models are ordered by Entry Top-1, and full reconstruction requires the canonical message, operational intent, and top-ranked entry to be correct. All six variants remain in the evaluation, and model-output failures remain in the denominator.

Entry Top-1 ranges from 35.19\% to 95.22\%. GPT-5.4 leads Entry Top-1, whereas GPT-5.6 SOL leads full reconstruction and has the lowest CER. Kimi K3 combines low CER with a lower Entry Top-1, confirming that character-level recovery cannot replace a separate evaluation of actionable-entry recovery.

Full stratified results by obfuscation group and all resampling intervals appear in the supplementary material.

\paragraph{Evidence-grounded web investigation.}
The web-only view evaluates exploration, evidence acquisition, and final judgment under correct website binding. Task correctness in the lower panel of Table~\ref{tab:main-results} is scored directly against human website gold labels, while the fixed multimodal evidence judge evaluates only the evidential relation between the conclusion and the observed trajectory. Runs without valid decisions remain failures.

Task~2 decision accuracy ranges from 26.3\% to 62.8\%, and hierarchical exact match ranges from 26.2\% to 61.0\%. GPT-5.6 SOL leads both metrics and decision macro-F1; Kimi K2.5 leads type macro-F1.

Evidence scores remain coverage-conditioned and do not define an overall ranking; complete coverage and four-dimensional results appear in the supplement.

\paragraph{Entry-gated end-to-end composition.}
Figure~\ref{fig:gated-e2e-overview} and Table~\ref{tab:gated-results} apply the fixed primary-entry gate to the same frozen web results. Gated accuracy ranges from 16.7\% to 60.8\%, with losses of 0.7--32.3 percentage points. The composition reuses each frozen Task~2 trajectory without a browser rerun.

\begin{table}[t]
    \centering
    \small
    \setlength{\tabcolsep}{4.2pt}
    \begin{tabular*}{\linewidth}{@{\extracolsep{\fill}}lrrr@{}}
        \toprule
        System & \multicolumn{1}{c}{Web-only} & \multicolumn{1}{c}{Gated} & \multicolumn{1}{c}{Loss}\\
        & Acc. $\uparrow$ & Acc. $\uparrow$ & (pp) $\downarrow$\\
        \midrule
        GPT-5.6 SOL & 62.8 & \textbf{60.8} & 2.0 \\
        GPT-5.4 & 53.7 & 52.3 & 1.3 \\
        Claude Opus 4.8 & 49.8 & 49.2 & 0.7 \\
        GPT-5.2 & 57.3 & 49.0 & 8.3 \\
        Kimi K3 & 47.5 & 44.5 & 3.0 \\
        Kimi K2.5 & 55.2 & 25.7 & 29.5 \\
        Qwen3.6 Plus & 51.7 & 19.3 & 32.3 \\
        Claude Sonnet 5 & 26.3 & 19.2 & 7.2 \\
        Doubao Seed 2.0 & 29.5 & 18.2 & 11.3 \\
        Kimi K2.6 & 41.5 & 16.7 & 24.8 \\
        \bottomrule
    \end{tabular*}
    \caption{Correct-routing Task~2 accuracy and entry-gated end-to-end accuracy (\%). All values use the same 600 websites per model. Loss is the difference in percentage points; 95\% intervals appear in the supplementary material.}
    \label{tab:gated-results}
\end{table}

\subsection{Diagnostic Analysis}

We assign each run to its first failed stage: execution, missing report, decision, or type; runs with both a correct decision and type are hierarchical successes.

\begin{figure}[t]
    \centering
    \includegraphics[width=\linewidth]{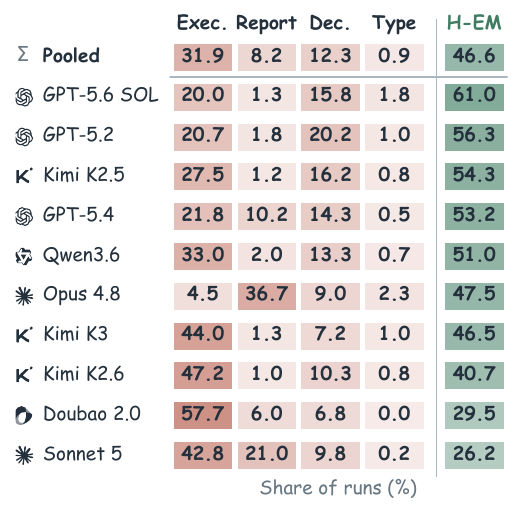}
    \caption{Task~2 first-failure attribution over all 600 runs per model; H-EM denotes hierarchical success.}
    \label{fig:failure-heatmap}
\end{figure}

Figure~\ref{fig:failure-heatmap} shows that execution is the largest pooled failure source at 31.9\%; only 0.9\% fail at typing after a correct decision.

\section{Conclusion}

\textsc{RiskChainBench} links obfuscated-message restoration and evidence-grounded web investigation in a resettable local sandbox. Across the same ten models, actionable-entry recovery and full reconstruction rank models differently, as do website decisions and fine-grained violation typing. The offline entry gate exposes upstream loss without rerunning website investigation; frequent execution failures show that reliable exploration remains prerequisite to evidence-grounded judgment.

The results use one trajectory per model--website pair, sparsely populate several violation types, and condition evidence diagnostics on successful judge coverage. Future work should measure rerun variance, broaden underrepresented risk types and languages, and validate the evidence rubric through a frozen human trajectory audit within the same isolated environments.

\section{Ethical Statement}

Synthetic, non-routable entries and isolated environments avoid contact with live services. Local substitutes replace personal identities, payment flows, account operations, and communication with third parties. Evidence packages retain only sanitized observations needed for evaluation. Benchmark outputs do not establish legal attribution.

\newpage
\bibliography{aaai2027}
\end{document}